\documentclass[conference]{IEEEtran}
\usepackage{cite}
\usepackage{amsmath,amssymb,amsfonts}
\usepackage{algorithmic}
\usepackage{graphicx}
\usepackage{textcomp}
\usepackage{enumitem}
\usepackage{xcolor}
\usepackage{csquotes}
\def\BibTeX{{\rm B\kern-.05em{\sc i\kern-.025em b}\kern-.08em
		T\kern-.1667em\lower.7ex\hbox{E}\kern-.125emX}}
\begin{document}
\pagestyle{plain}

\title{Internet of Robotic Things: Current Technologies, Applications, Challenges and Future Directions}

\author{\IEEEauthorblockN{Davide Villa, Xinchao Song, Matthew Heim, Liangshe Li}
	\IEEEauthorblockA{\textit{Department of Electrical and Computer Engineering} \\
		\textit{Northeastern University}\\
		Boston, MA, 02115 USA \\
		\{villa.d, song.xin, heim.m, li.liangs\}@northeastern.edu}
}

\maketitle

\begin{abstract}
	Nowadays, the Internet of Things (IoT) concept is gaining more and more notoriety bringing the number of connected devices to reach the order of billion units. Its smart technology is influencing the research and developments of advanced solutions in many areas. This paper focuses on the merger between the IoT and robotics named the Internet of Robotic Things (IoRT). Allowing robotic systems to communicate over the internet at a minimal cost is an important technological opportunity. Robots can use the cloud to improve the overall performance and for offloading demanding tasks. Since communicating to the cloud results in latency, data loss, and energy loss, finding efficient techniques is a concern that can be addressed with current machine learning methodologies. Moreover, the use of robotic generates ethical and regulation questions that should be answered for a proper coexistence between humans and robots. This paper aims at providing a better understanding of the new concept of IoRT with its benefits and limitations, as well as guidelines and directions for future research and studies.
\end{abstract}

\section{Introduction}\label{sec-intro}

Over the past decades till today, the Internet of Things is experiencing an exponential growth and attention \cite{bib1} \cite{bib2}. Its smart technology allows the creation of a network of real-world objects, namely ‘Things’, with the purpose of connecting everything through the ability of exchanging information over the Internet \cite{bib14} \cite{bib15}.
This revolution is shaping all aspects of the human life.
The IoT concept is being adopted by more and more organizations in several fields, e.g. military, robotics, healthcare, nanotechnology or space, creating the Internet of X Things, where X is the relevant area. Currently, according to the analyst firm Gartner \cite{bib31}, 127 new IoT devices are connected to the Web every second bringing its overall number to reach 75 billion units by 2025. The global spending on the IoT should reach 1.29 trillion dollars during 2020 with a 93\% of adoption of IoT technology by the enterprises.

An area where the Internet of Things technology is finding fertile ground is the robotics. Robotics consists of a modern and fast-evolving technology that is bringing enormous changes in several aspects of human society over the past decades. It can be defined as "The branch of engineering that involves the conception, design, manufacture and operation of robots" \cite{bib32}. Its application can range from the execution of repetitive and tedious jobs in a manufacturing line to helping and performing critical or dangerous tasks unaffordable by a human being, such as rescuing in disaster areas or extra-terrestrial operations. The robots initially used in these applications were just single machines with limitation given by their hardware components and computational abilities. To resolve those issues, the robots were started to be connected in a communication network through a wired or wireless creating a Networked Robotic System \cite{bib33}. However, they suffered from inherent resource constraints which led to network latency, limited memory and low computational and learning capabilities. Today real-life scenarios demand fast and complex task executions which require sophisticated data analysis and high computational abilities. The solution to these limitations have been recently tried to be addressed in the novel form of Cloud Robotics \cite{bib12} which takes advantage of a cloud infrastructure to access resources on-demand and to support the operations. However, these systems are affected by new issues like interoperability, network latency, security, quality of service and standardization \cite{bib34}. The Internet of Robotic Things perfectly fits in this scenario aiming to overcome these constraints by combining together the IoT with robotics and leading to a more efficient, smart, adaptive and also cheaper robotic network solution.

This paper has the aim to provide a comprehensive description of the Internet of Robotic Things concept showing its general architecture, benefits and challenges as well as some practical use cases and guidelines in the hope of giving awareness on this subject and inspiring future research. The rest of the paper is organized as follows. Section \ref{sec-overview} gives an overview of the IoRT concept, while Section \ref{sec-arch} explains in more details its architecture. Some practical use-case examples are discussed in Section \ref{sec-domain}. Section \ref{sec-challenge} addresses the main challenges that the current research is facing during the development of an IoRT system. Moreover, Section \ref{sec-ethic} introduces the ethical issues that arise in the use of IoRT in the everyday life, while Section \ref{sec-regul} and Section \ref{sec-future} discuss the topics of regulations and how IoRT can shape the modern society. Finally, the conclusions are addressed in Section \ref{sec-conclusion}.

\section{Overview of the Internet of Robotic Things}\label{sec-overview}

The Internet of Robotic Things (IoRT) is the merge between IoT and robotics. The term IoRT was coined by Dan Kara in a report of ABI Research \cite{bib35} to denote intelligent devices that can monitor events, gather data from a variety of sources and sensors, and exploit both a local and a distributed intelligence to control objects and determine the best action to follow. More in general, the IoRT can be seen as a global infrastructure enabling advanced robotic services thanks to the interconnection of robotic things. The robotic things can exploit the benefits of modern communication and interoperable technologies based on the cloud, through the IP protocol and its IPv6 version, to take advantages regarding data processing, memory storage, computational overhead and security. This mitigates and can resolve the previous robotics issues. Moreover, the Internet of Robotic Things goes beyond the networked and cloud robotic paradigms by taking advantage of the enabling IoT technologies. This aims at integrating heterogeneous intelligent devices empowering enormous flexibility in the design and implementation of a distributed architecture which provides computing resources both in the cloud and at the edge.

The IoRT multidisciplinary nature follows the same trend of the IoT by providing advanced robotic capabilities leading to the rise of interdisciplinary solutions for various and different disciplines.
From a technological perspective, IoRT enhances the ordinary robot abilities which are usually classified in 3 groups: basic (perception, motion, manipulation); higher-level (decisional, autonomy, interaction cognitive); and system-level (configurability, adaptability, dependability). To achieve this, in addition to the IoT features, IoRT solutions are supported by other technologies such as: multi-radio access to link smart devices together; artificial intelligence to generate optimized solutions for complex problems; and cognitive technologies to allow operational efficiency.

To summarize, the Internet of Robotic Things is designed to be placed at the summit of the Cloud robotics and combines the IoT technology features with the autonomous and self-learning behavior of connected robotic things in order to generate advanced and smart solutions making an optimal use of distributed resources.

\section{IoRT architecture}\label{sec-arch}

\begin{figure}[t]
	\vspace*{-4.5mm}
	\centerline{\includegraphics[trim=10cm 2.8cm 10cm 2.4cm,clip,width=8cm]{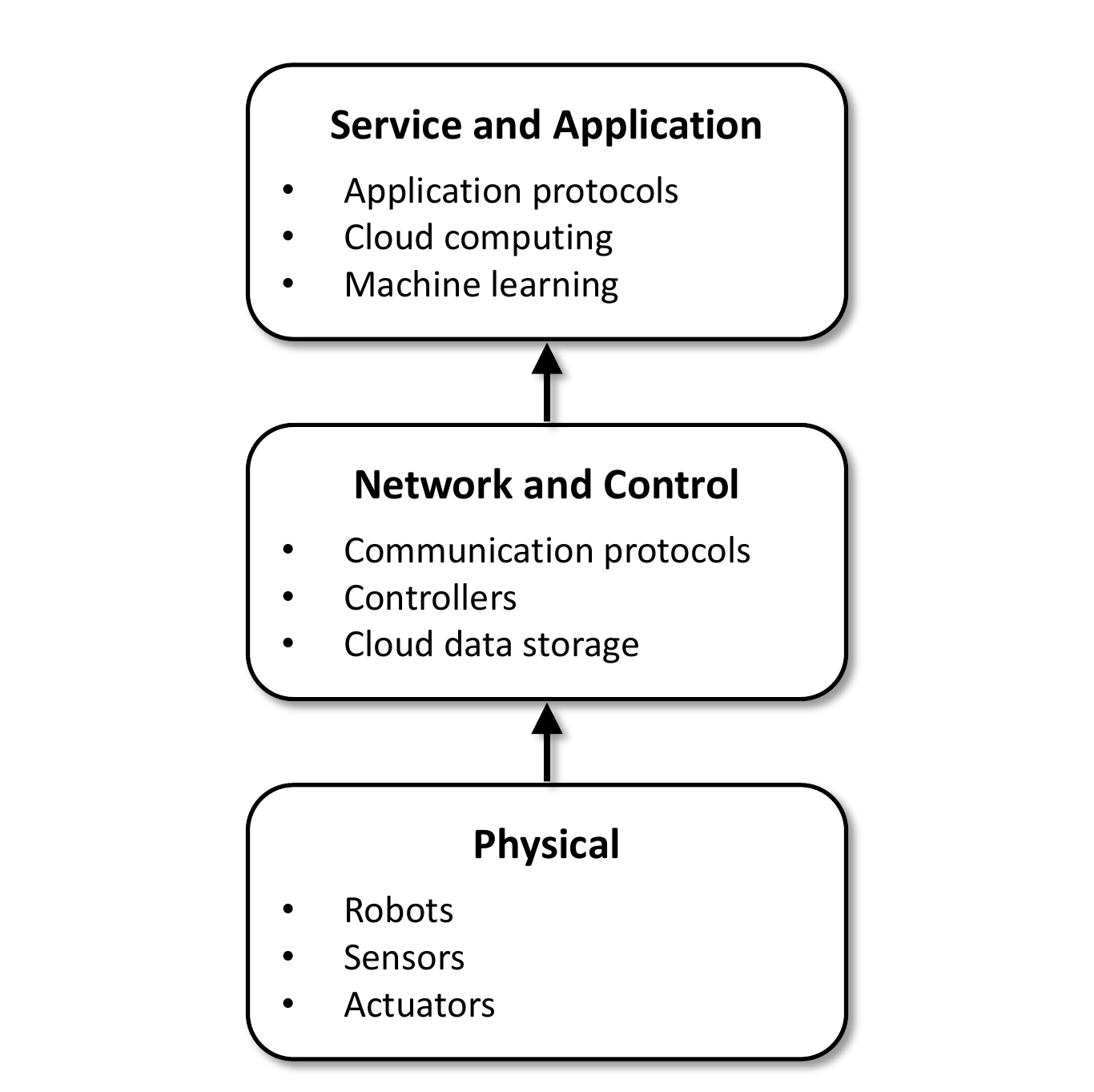}}
	\caption{Internet of Robotic Things architecture overview.}
	\label{fig:iortarchitecture}
\end{figure}

The architecture of an Internet of Robotic Things system can be described as being composed by three main layers: Physical, Network and Control, Service and Application (Figure \ref{fig:iortarchitecture}) \cite{bib3} \cite{bib4}. Its composition recalls the typical structure of the OSI model but with a different perspective which takes into account also the robotic part of IoRT.

\subsection{Physical}

The physical layer consists of the lowest part of the IoRT architecture. It includes everything related to the hardware components, such as robots, sensors and actuators. The robots can be described as intelligent agents that can communicate between each other and have the ability to establish a more complex system in order to accomplish a specific objective thanks to a sequence of distributed actions. A robot can be represented for example with a vehicle, a drone, a home appliances, or a health care equipment. On the other hand, sensors refer to any kind of system that the agent can use to perceive a data from the surrounding environment, while actuators to any kind of tool used to perform an action. Both sensors and actuators can also be integrated inside a single robot system to enhance and optimize the performance. As in the standard OSI model, the purpose of the physical layer is to abstract all of technical information and provide a set of services about its periphery to the network and control layer.

\subsection{Network and Control}

The Network and Control layer provides an efficient integration of the hardware components of the physical layer by means of control processes and communication protocols. Robotic devices are able to send information to other robots or to the cloud using various methods. There are a few main protocols used for communication depending on the range between sender and receiver and the power required to send data over that range. Using any protocol should allow data to be communicated over the Internet. At the link and physical layers, common short-range protocols are IEEE 802.11 (Wi-Fi) and Bluetooth. Common long-range protocols are Sigfox, LoRaWAN, and NB-IoT. Another protocol worth noting is IEEE 802.15.4 which makes the basis for 6LowPAN and Zigbee. At the network layer, common protocols used for IoT are 6LowPAN and the Routing Protocol for Low Power and Lossy Networks (RPL). Table \ref{tab:iortprotocolstack} shows applicable protocols for robotics at different layers of the protocol stack.

\begin{table}[t]	
	\caption{IoRT protocol stack}
	\def\arraystretch{1.4}
	\begin{center}
		\begin{tabular}{|p{1.5cm}|p{3.4cm}|p{2.6cm}|}
			\hline
			\textbf{Layer}       & \textbf{IoRT protocols}                          & \textbf{Usage}                                               \\ \hline
			Application & CoAP, MQTT, REST API, XMPP, TLS                       & Data streaming                                      \\ \hline
			Transport   & UDP, TCP                                              & End-to-end communication                            \\ \hline
			Network     & IPv6                                                  & Internet connection                                 \\ \hline
			Link        & Wi-Fi, 802.15.4, LoRaWAN, NB-IoT, Bluetooth, Sigfox & Robot-to-Robot, Access Point, gateway communication \\ \hline
			Physical    & Robots, Sensors, Actuators                            & Data acquisition                                    \\ \hline
		\end{tabular}
		\label{tab:iortprotocolstack}
	\end{center}
	\vspace*{-4.5mm}
\end{table}

6LoWPAN allows a Low Rate Wireless Personal Area Network (LR-WPAN) to use IPv6. This protocol is particularly used in the IoT for its ability to be low-energy and low-cost which is crucial for small, battery-powered robots or sensors. Protocols that were initially designed for wired networks, such as Precision Time Protocol (PTP), can be adapted for 6LoWPAN \cite{bib27}. RPL, also using IPv6, creates a directed acyclic graph to structure the network and allows three types of communication: Multi-Point-to-Point, Point-to-Multi-Point, and Point-to-Point.

Sigfox, a low-power wide-area network (LPWAN), works with Ultra Narrowband (UNB) signals at frequencies lower than 1 GHz with a very low datarate of up to 100 bits/sec \cite{bib43}. The result is a very low energy signal that can pass through buildings and reach underground. This technology could be effective for robots to send short status messages from a long distance, but they must talk through a mobile gateway first.

LoRaWAN is another LPWAN that uses chirp spread spectrum (CSS) around the same frequencies of Sigfox. However, the two technologies do not interfere with one another since the power of LoRaWAN signals are spread over a band of frequencies. LoRaWAN uses different spreading factors that affect the datarate and possible range of transmission. With spreading factors 7 to 12, each one is orthogonal to the others.

While Sigfox and LoRaWAN works on unlicensed frequency bands, NB-IoT works on a licensed spectrum which guarantees more reliability with less interference. In an analysis of offshore unmanned aerial vehicle (UAV) monitoring cargo vessels, NB-IoT base stations (BS) could be used effectively on UAVs acting as relays \cite{bib44}. Localization for UAVs is an ongoing research area addressing problems such as network failure when losing line of sight with the UAV and GPS limitations.

Security can be addressed at all levels of the protocol stack: application, transport, network, link, and physical. Covering the application and transport layers, the Transport Layer Security (TLS) protocol can provide secure and trusted data transmission by applying message integrity, authentication, and confidentiality. At the network layer, RPL can be used in pre-installed and authenticated modes for message integrity and secure communication, respectively. At the link and physical layers, Wi-Fi can also use authentication and data encryption. The use of frequency hopping and spread spectrum can alleviate the problems of interference and jamming. However, these examples only cover a small subset of the available protocols, so security is an ongoing critical issue with IoT devices.

Many of the application areas for the IoRT, such as manufacturing, are high-stakes environments and therefore should have cybersecurity measures to protect the network from malicious intent. One way to address the problem of malicious nodes in a network is trustworthiness management \cite{bib28}. There may be subjective trustworthiness, where every node in a network computes their own trust of other nodes. In contrast, there may be objective trustworthiness, where the trustworthiness of a node lies in a distributed system that is shared among the nodes in the network. Subjective trustworthiness can slow a network down but can be more protected from malicious intent than an objective trustworthiness system.

Looking to the future, one area of research that shows promise in the IoRT is Large Intelligent Surfaces (LIS) \cite{bib40}. Walls can have a large number of antennas and act as a massive multi-input multi-output (MIMO) system. This technology could allow for more capabilities in a smart environment by making the infrastructure more active. Robots could receive data from large intelligent surfaces and could even interact with them in the case of a wall-climbing robot.

\subsection{Service and Application}

The Service and Application layer resides at the top of the IoRT architecture and relies on program implementations to properly perform the end-user operations of monitoring, processing and controlling. There are several options for protocols. One example is CoAP (Constrained Application Protocol) which uses UDP at the transport layer. It is useful for resource-constrained devices on low-power, lossy networks. The security and applicability of using CoAP for IoT devices is questionable. A more widely used protocol is MQTT (Message Queue Telemetry Transport) which uses a publish/subscribe architecture. MQTT aims to be lightweight and reliable over unreliable networks. A commonly used communication architecture in the IoT is REST, or REpresentational State Transfer. It only relies on simple HTTP methods (GET, POST, PUT) and its ubiquity is a strength. However, REST adds to the complexity of a robotic system and may be avoided. Another common application protocol is XMPP (eXtensible Message Presence Protocol). It is an open standard that aims to send close to real-time messages between devices in a decentralized manner. As a case study, a cloud platform for an exoskeleton robot chose to use MQTT to transmit data to the cloud and HTTP to communicate from the cloud \cite{bib45}.

An integral part of current robotics is machine learning techniques that can be used for tasks such as mapping, localization (knowledge of a robot’s own location), and learning the environment. Machine learning has traditionally been used for object detection, tracking, and classification. However, its applicability to robotic learning means it is continually adapting field being used to solve more complex problems. Cloud robotics is used to keep robotic things operating at a low power and intelligence level while having a powerful data center to communicate with. Robots may send data to the cloud for mapping and localization, perception, and actuation purposes. The cloud can be accessed for more data that robots may need for training neural networks. The robot can also learn by taking models from the cloud without having to perform extra computations. The cloud may provide more general knowledge while the robot learns its local environment. Cloud robotics also allows for human-robot interaction by having humans make decisions for robots or control robots over a network. Having multiple robots communicate over a network simultaneously can result in high latency and data loss. There may be much more data that robots need to send to the cloud or to other robots. For example, streaming LIDAR over Wi-Fi can result in a network failure with multiple robots \cite{bib29}. Robots can use deep reinforcement learning (RL) to query the cloud intelligently with a minimal frequency and associated cost.

It is important for a robot that comes in contact with humans to be able to act and react with humans differently than other robots. To accomplish this, a robot must express personality and connect with the emotions and storytelling of humans so that humans can understand what the robot is doing and for what reason. A robot whose behavior is influenced by a robot can be called a socially responsive robot \cite{bib30}. A promising technique is when a human can teach a robot the movements of a behavior by showing an example. RL can be used by the robot to learn the actions and states that make up a behavior, such as dancing, without understanding the dynamics of systems. Effective application areas of socially responsive robots include hospitals, rehabilitation, and generally smart environments or indoor areas with many types of sensors.

\section{Application domains}\label{sec-domain}
IoRT has been widely used in many fields and the machine learning concept is undoubtedly one of the key links. In particular, robots combined with IoT can bring lots of benefits in doing repetitive tasks and in operating in disaster areas where it is unaffordable for a human being to work properly.

\subsection{Disaster response}

\begin{figure}[t]
	\centerline{\includegraphics[trim=1.7cm 3.8cm 1.7cm 3.8cm,clip,width=9cm]{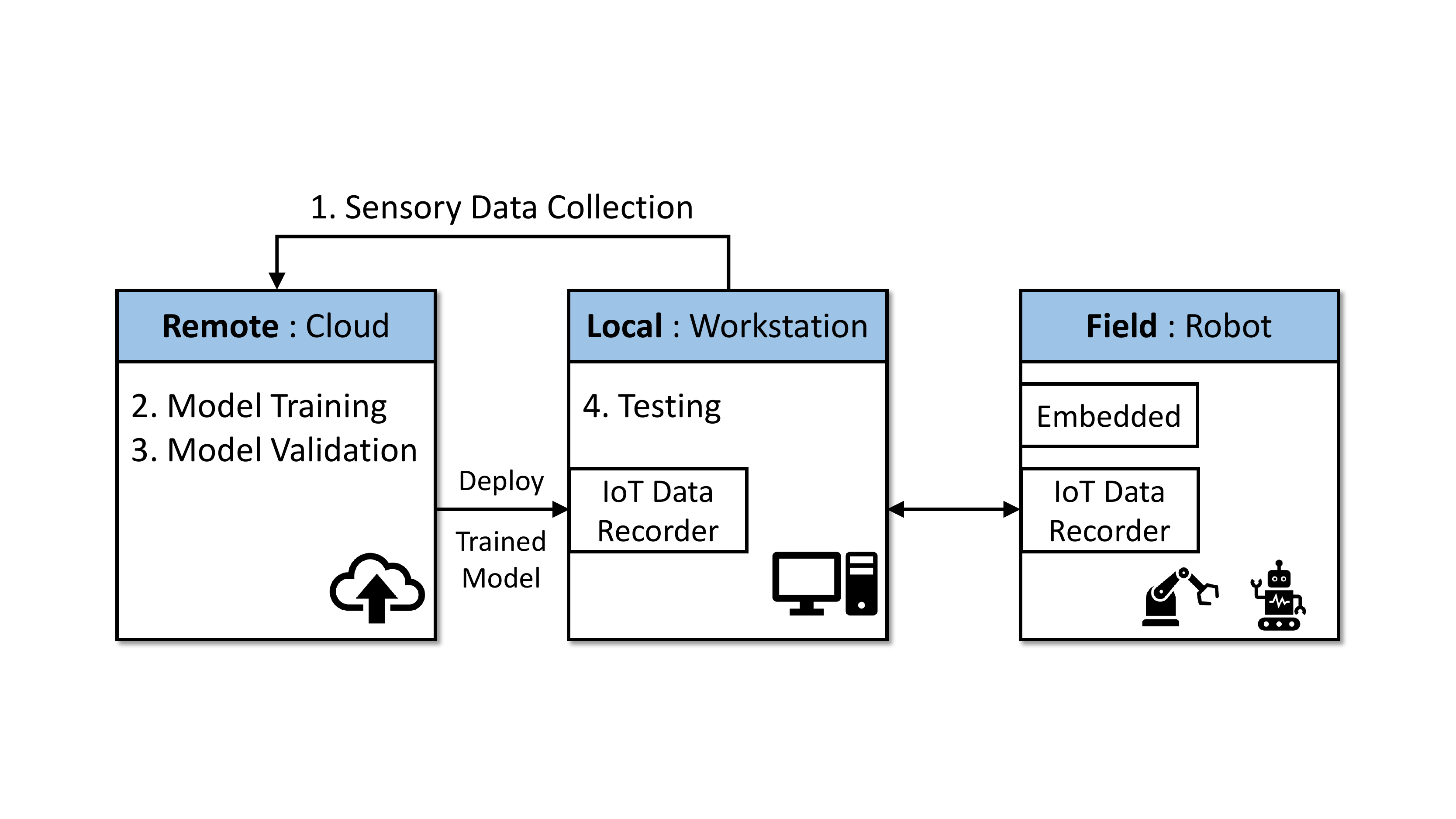}}
	\caption{IoRT in disaster response use case example.}
	\label{fig:disasterresponse}
	\vspace*{-4.5mm}
\end{figure}

In \cite{bib23} the authors exploit an IoRT architecture for a disaster response. A lot of major disasters happen on Earth every year, e.g. earthquakes, tsunamis or typhoons. When a calamity of this magnitude happens, the time is very precious. The most urgent task to achieve is to save as many people as possible in the shortest time. Robots can offer a lot of help in this kind of situations after they have received a deep learning training. In order to accomplish this, the first step is to use mobile robots to collect data from the local environment. A remote network elaborates these data to build an AI model which is evaluated by the internal system. The model is also distributed from the cloud to the local workstation for further performance testing. After it has reached a certain level of reliability, the model is ultimately deployed into the robots for the next level of learning process. The proposed architecture can be schematically represented in Figure \ref{fig:disasterresponse}.

This solution can make the robots have enough experience on how to respond to disasters in one area. However, different places might present different environments. If the robot has information about the environment of only one single type of place, it cannot be appointed to other places to carry out rescue missions. A solution to this issue is to consider to store in the cloud different AI models for various environments. Once a disaster happens, the AI model of that environment can be directly deployed in the robots so that they are immediately able to successfully accomplish the rescue mission in that area.

As an example, in the event of an earthquake, the robots might be able to detect the victim’s position thanks to their sensors and to help on removing reinforced concrete that might have collapsed. Additionally, If a victim is trapped under the ruins and is not able to get out momentarily, the robots can be responsible for transporting food and water to help the victim maintain the nutrition they need. The overall success rate of rescue operations and the survival rate of victims can be greatly improved thanks to the use of these robots.

\subsection{Precision agriculture}

\begin{figure}[b]
	\centerline{\includegraphics[trim=6cm 1.8cm 6cm 1.8cm,clip,width=8.6cm]{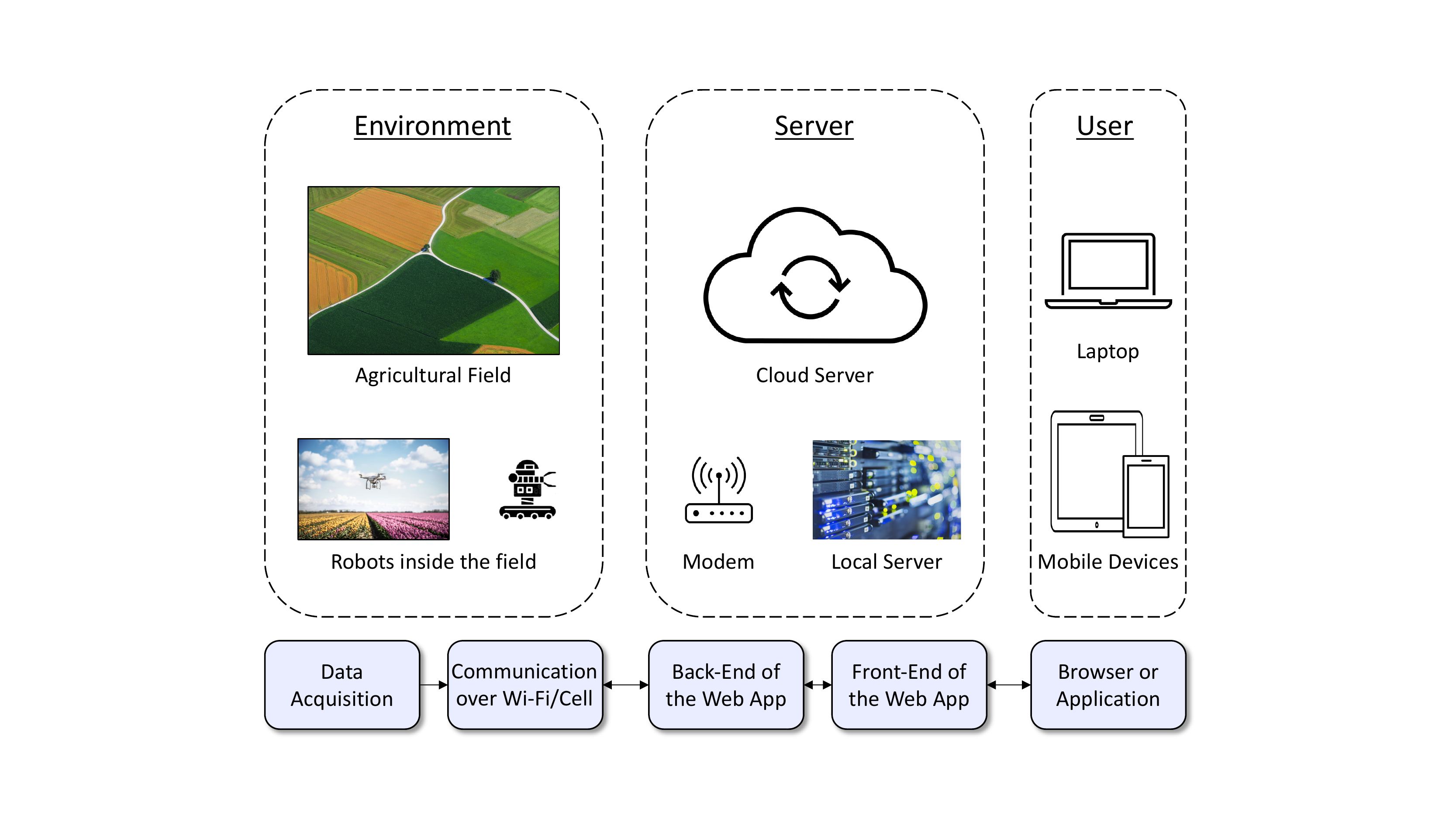}}
	\caption{Conceptual design of the agriculture system.}
	\label{fig:agricutlure}
\end{figure}

Authors in \cite{bib16} provide an IoRT application in the field of precision agriculture. Figure \ref{fig:agricutlure} presents a sketch of an agricultural IoRT system. First of all, the mobile robots in the environment collect data using on-board sensors. The data contain information about temperature, humidity, pressure and light measurements. The server can retrieve these data from the mobile robots using cellular communication or Wi-Fi in order to process them. After that, the data will be deployed to web applications and the user is able to visualize them through a web interface.

This solution brings the advantages of the IoRT into full play.
The mobile robots are able to collect data on-demand instead of the human. These processed data can be used to develop appropriate agricultural plans during climate changes or hazard situations. This will result in a saving of time and resources as well as an increasing of the productivity.

However, static robots or devices are also needed in this system to measure weather conditions or provide alarm signals in case of hazards. An appropriate website design is a key point to present information handily and to visualize data in a more apprehensive way to the user. Further work can also focus on the subsequent application of the processed data. For example, the users could assign further instructions to mobile robots through web applications after the data acquired have been collected and analyzed.

\subsection{Medical services}

Another field where IoRT can bring benefits is the medical one. In \cite{bib22} authors have developed an application comprising a massage robot. The robot exploits an IoT architecture with sensors to collect data real-time and combine them with a smart control system to apply the proper pressures of the desired massage technique. The robot is a mixture of data-mining and brain-based networks. This use case perfectly represents the current trend of the technology showing how the IoRT is fitting in the medical health area. This opens the doors for a huge variety of future development research.

\subsection{Smart Restaurant}

\begin{figure}[b]
	\centerline{\includegraphics[trim=1.7cm 4.1cm 1.7cm 4.1cm,clip,width=9.1cm]{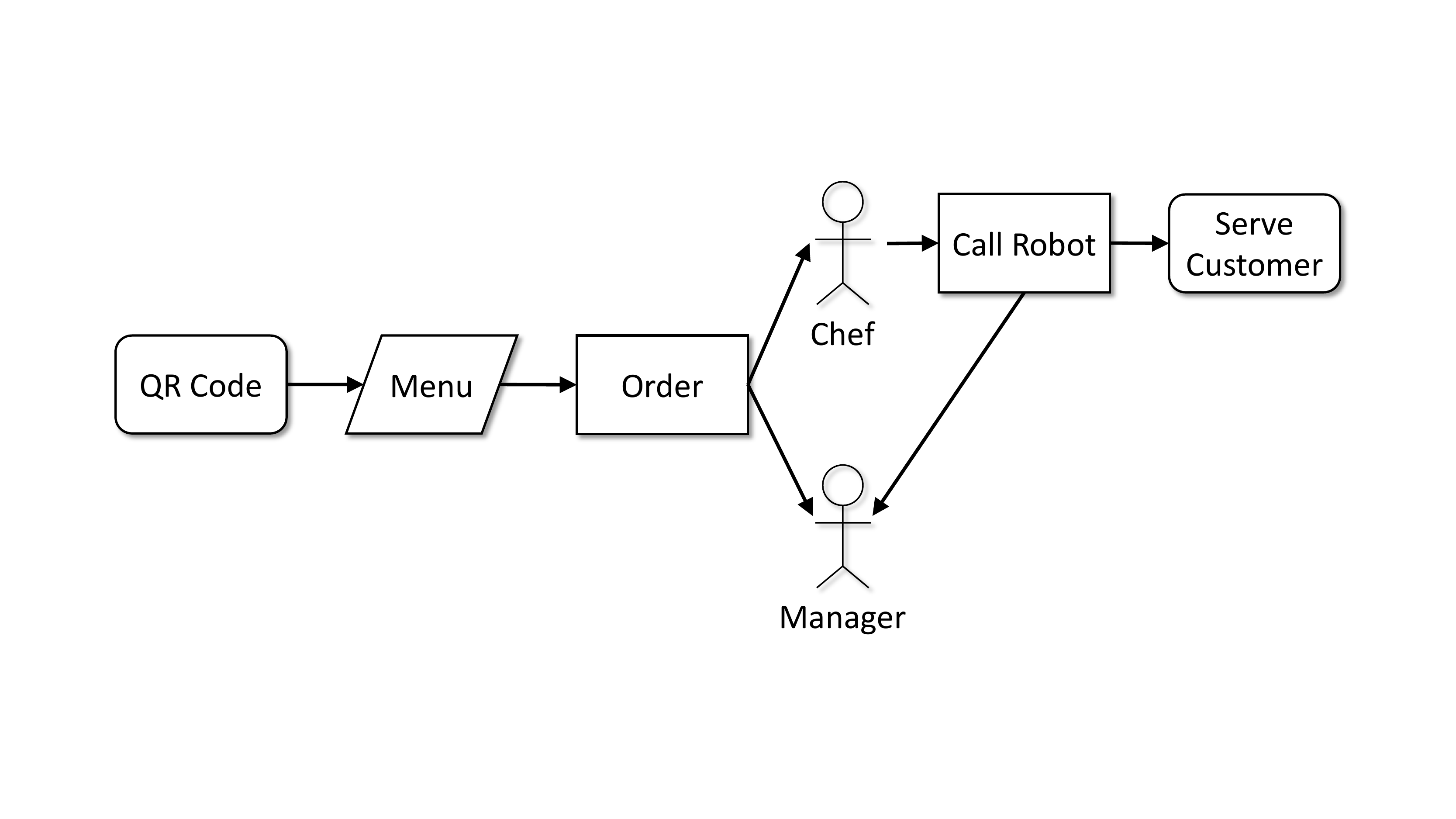}}
	\caption{Main blocks of the waiter robot system.}
	\label{fig:waiterrobot}
\end{figure}

The revolution in the use of IoRT is shaping all aspects in the people's daily life. In \cite{bib36} a robot acting as a restaurant waiter is proposed. Figure \ref{fig:waiterrobot} shows the main blocks of the system. The information about table number and menu are stored in a QR code. The orders are managed by a website and received by both the cuisine and the manager together with the table number. As soon as the orders are ready, the chef utilizes the robot to deliver them to the customers. The whole system is automatic and the information are stored in the cloud.

The solution is simple and feasible. However, some key problems still need to be addressed. For example, the robot lacks some emergency measure and it may have issues if the walking route is blocked. Another significant concern is that the waiter robot is not able to speak with customers, which may leave a bad impression at the end. Nevertheless, this IoRT solution has the potential to significantly increase the employer profit by reducing the personnel expenses and to provide a faster service to the customers.

In short, the IoRT applications are being largely considered in the academic field and some solutions have been already used to develop a commercial product. With the imminent introduction of 5G and, in the future, of 6G, the IoRT technology has the potential to increase its performance and to continue to be considered a valuable field of research.

\section{Technology Challenges}\label{sec-challenge}

Although growing rapidly, IoRT is still a novel area in its preliminary stage and is facing many challenges that need to be overcome. This section introduces two of the most significant challenges, i.e. data processing and security, and discusses possible solutions to these problems.

\subsection{Data Processing}

Because of the dual cyber-physical nature of IoRT, huge amounts of data are collected in various forms and from different sources, e.g. data from various types of sensors or information for robot control. These data need to be transmitted from the nodes to the cloud where they can be remotely processed. This requires a higher capability of data processing than a regular IoT application, and consequently may cause the following challenges.

\begin{itemize}
	\item Massive data streaming and processing usually demand a significant communication bandwidth and a powerful computing capability in an IoRT network. As an example, a group of security robots may need to send different visual and acoustic data at the same time during their patrol. This requires fast data processing to make sure the system can rapidly respond to possible incidents.

	\item Latency on real-time applications has a considerable impact on the operating efficiency and safety. Since many IoRT systems, such as a robotic transpiration system in a smart factory, are actively interacting with the environments in a time-sensitive manner, undesired latency could cause operation failure or even accidents.

	\item Unstable communication in a complex environment may significantly reduce the system performance. For example, a medical IoRT system in a hospital may suffer from the hospital building's complex indoor space and interference from other medical devices such as Magnetic Resonance Imaging (MRI).
\end{itemize}

To address those issues, possible solutions may include using different computing frameworks, new communication technologies, and optimization.

Edge computing and fog computing are two of the latest popular solutions towards computational challenges. Both architectures aim at distributing the data computation and storage all over the network, so that part of the data processing can be done closer to the end nodes. The difference between them is that edge computing is trying to process more data locally, while fog computing is using IoT gateways to do that. IoRT systems might exploit one or both technologies based on the working environment and requirement specifications \cite{bib7}.

As for communication, heterogeneous IoT technologies may be applied. It represents a multifaceted network integrating different communication architectures, including Wi-Fi, cellular networks, and Wireless Sensor Networks (WSN). It can provide more choices and flexibility for massive data exchanges among different types of nodes \cite{bib15}.

Another possible measure is to use higher-speed wireless communication, such as Terahertz communication. Terahertz is the next generation of wireless communications which aims to use terahertz band (0.3 \textit{THz} - 10 \textit{THz}) to unlock significantly wider bandwidth and achieve higher communication speed estimated to exceed 100 \textit{Gbit/s}. It will be fast enough in order to allow massive data exchange in most common IoRT networks.

Finally, good optimization frameworks, which take pre-processing data, computation modes, and communication technologies together into consideration, can help to obtain the optimal computation strategies making the whole IoRT more efficient \cite{bib7}.

\subsection{Security and Safety}

Security is a crucial issue in this area. Many application domains of IoRT have high demands on safety and stability among their operations. Security is also a challenging problem since it is related not only to the security of the Internet of Things but also to the safety of the robots. As we discussed above, there are massive amounts of various data in the IoRT systems that need to be exchanged between the robots and the cloud as well as between robot and robot. This leaves more room for security failures, from leaking sensitive data to cyber-attacks. Some typical problems are listed below \cite{bib3} \cite{bib37}.

\begin{itemize}
	\item Insecure communication between cloud and robot or between robot and robot may cause sensitive data to be exposed or hacked. The less the communication is secured, the easier a hacker breaks in, especially for those professional hackers sponsored by business competitors or even hostile foreign governments.

	\item Authentication failure may allow hackers or unauthorized personnel to easily obtain access to the system, hijacking its parts or even the whole architecture, and sabotaging the robots by reprogramming them.

	\item Robot failures, no matter whether caused by unexpected hardware faults or intentionally cyber-attacks, may damage the environment, injure people, or even cause secondary accidents.
\end{itemize}

Those issues can be addressed and mitigated by the following measures. The first one is to use encrypted communication based on protocols like TLS and SSH to secure communication. New technologies like quantum communication may also be a future direction to provide a possible once and for all solution to that. Additionally, strengthening authentication by using different measures such as data fragments, externalized authorization, or granular permission, can improve the authentication security \cite{bib38}. Some new technologies like block-chain could also provide further assistance. Last but not least, safety measures on robots, such as hardware failure monitoring, emergency detection and stop, system redundancy and backups, could help to guarantee the safety and robustness of the robotic part of the IoRT architecture.

\section{Ethical Issues}\label{sec-ethic}

As an integration of IoT and Robotics, IoRT is also facing ethical issues and challenges from both sides. There are three main ethical issues that should be taken into consideration: robot ethic, property rights of information, and data privacy. It is crucial to address those issues properly for the harmonious coexistence between robots and humans in the modern society.

\subsection{Robot Ethic}

The ethic of a robot system is a topic that has been discussed since the dawn of the robot concept. The famous \textit{Three Laws of Robotics} by Isaac Asimov state \cite{bib41}:

\begin{enumerate}
	\item \textit{A robot may not injure a human being or, through inaction, allow a human being to come to harm.}
	\item \textit{A robot must obey orders given it by human beings except where such orders would conflict with the First Law.}
	\item \textit{A robot must protect its own existence as long as such protection does not conflict with the First or Second Law.}
\end{enumerate}

The core idea here is clear: robots should benefit but not harm users. However, as a technology with ethical implications, IoRT cannot achieve this goal by itself but needs the relevant society, including robot manufacturers, users, and governments, to take actions on effective policies and novel practices to prevent the misuse of robots.

\subsection{Property Rights}

One of the ethical questions on data faced in the IoRT area regards the origin of these information and the owner of them. Since artificial intelligence and robots are involved and generate intelligent information, the answer could be even more tricky than for common IoT applications. As an example, one of the difficulties is how to properly identify the authors of the data \cite{bib42}, which may not be clear in many cases.

\subsection{Privacy}

Another possible ethical question can be who should access the data and how these information should be used. Especially for sensitive data, such as medical or financial data, things get more serious. According to the current society consensus on privacy, no data collector should have unlimited and unregulated rights to access or use any data without proper permission or authorization. Due to the high capability of most IoRT applications to collect data from both their working environment and interacted clients, this issue needs to be addressed carefully.

\section{Regulations}\label{sec-regul}

\begin{figure}[t]
	\centerline{\includegraphics[trim=8.5cm 9.3cm 8.5cm 6.1cm,clip,width=8cm]{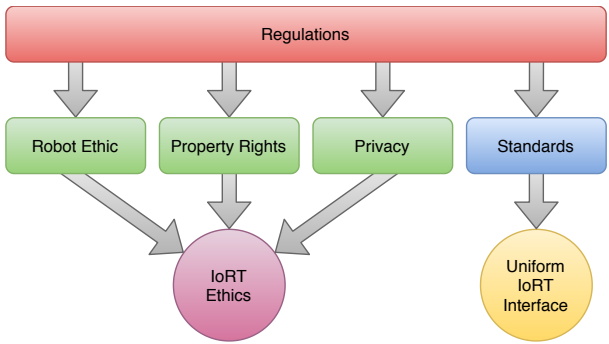}}
	\caption{The relationship between regulations and IoRT ethics and technologies.}
	\label{fig:regulations}
	\vspace*{-4.5mm}
\end{figure}

From the above discussion, the necessity to have regulations for the IoRT area can be clearly realized \cite{bib11}. However, by 2019, there was still no clear regulatory framework specifically on IoRT, but only partial regulations that may have been applied on relevant technologies \cite{bib17}. As an example, in the European Union, robots can be regulated as general products under safety regulations Directive 2001/95/EC. Depending on their use, such as industrial robots or medical devices, they may also be subject to some specific legislation like Machinery Directive 2006/42/EC or Regulation 2017/745. Regarding data security, General Data Protection Regulation (GDPR) has regulated how to properly process sensitive data. However, since all those regulations are only for general robots or data alone, they cannot reflect and regulate the unique and specific needs in the area of IoRT which combine robots and IoT technologies together. In particular, due to the cyber-physical nature of IoRT systems, in many cases it is vague whether they should be classified as products or services. Since both of them are usually regulated separately, it is sometimes hard to determine the liability when an incident occurs. The negative impacts are not only on ethical issues but also on technologies. The absence of regulatory frameworks leads to a lack of well-defined standards or uniform IoRT technological interfaces, which brings significant difficulties to integrate together different IoT and robotic devices from various manufacturers, especially when they are highly divisive and heterogeneous \cite{bib4}.

The current situation could be improved by issuing new mandatory regulations from governments or voluntary standards from industrial communities, although this might create other challenges regarding, for example, how to clarify liability \cite{bib17}. As the usage of IoRT technologies becomes increasingly popular, this will be eventually solved through the collaboration of policymakers, companies, and consumers. Future regulations should improve ethics and build a uniform technological interface for IoRT as shown in Figure \ref{fig:regulations}.

\section{Future Vision}\label{sec-future}

\begin{figure}[t]
	\centerline{\includegraphics[trim=7cm 1.1cm 7cm 1.1cm,clip,width=9cm]{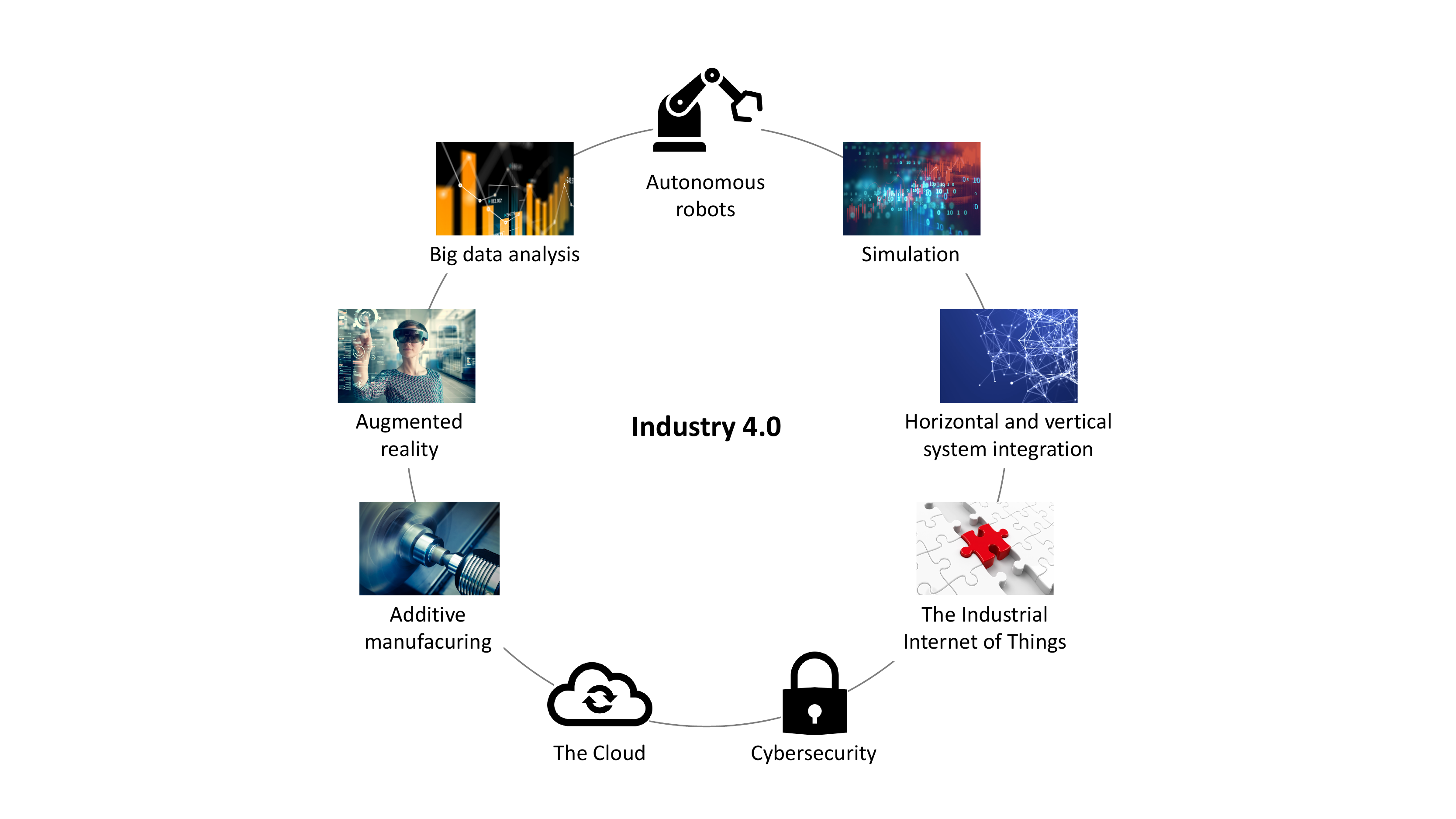}}
	\caption{Nine key technology in Industry 4.0 \cite{bib39}.}
	\label{fig:industry40}
\end{figure}

The Internet of Robotic Things seems to have all qualities and prerequisites to become one of the core technologies to support the future smart society. As an example, IoRT is the cornerstone for the imminent smart manufacturing \cite{bib5} and stands as one of the nine key technologies that form the foundation of Industry 4.0 \cite{bib39} (shown in Figure \ref{fig:industry40}). It may even lead to the realization of the so called "zero-labor factory" which aims to fully automate manufacturing.

Likewise, the integration of the two technologies of Internet of Things and robotics makes IoRT to perfectly fit in the current trend of the smart city\cite{bib46}, where massive sensors and robots are widely deployed and connected through IoT all across the urban area. This involves an automation of public facilities leading to more accessible and convenient services for all citizens.

\section{Conclusion}\label{sec-conclusion}

The Internet of Robotics Things is an emerging area aiming at integrating robotic technologies into IoT environments. Internet of Robotic Things enables robotic systems to communicate with each other, connect to the cloud, and share data and knowledge to accomplish sophisticated tasks. In order to provide a better understanding of IoRT and its current development directions, in this paper, we have reviewed its relevant concepts, technologies, applications, and current challenges, as well as provided a future vision of what the IoRT can lead to. We have first discussed the three main layers that form the IoRT architecture, i.e. physical, network and control, and service and applications. Then, we have presented four real use-case applications to show IoRT systems' capabilities and what they can bring to modern society. Next, several critical technological challenges, including data processing, security, and safety, have been discussed, and possible solutions have been proposed. We have also introduced some sensitive ethical issues that need to be properly addressed to ensure a successful co-existence between robots and humans. Moreover, the current lack of regulations may negatively impact the further development of IoRT and, for this reason, this should be solved through the efforts of the whole society. Finally, as we have presented in the last part of the paper, we consider the Internet of Robotic Things as a vital technology that has the potential to bring enormous advantages in modern society and that can contribute to the creation and development of the future smart manufacturing and smart cities.

\nocite{*}
\bibliographystyle{ieeetr}
\bibliography{./references}

\end{document}